# Causal Heterogeneous Graph Learning Method for Chronic Obstructive Pulmonary Disease Prediction

*Leming Zhou[1], Zuo Wang[2], zhigang Liu[3]\**

[1]*College of Computer Science and Technology,*
*Chongqing University of Posts and Telecommunications, Chongqing, China*
[2]*College of Computer and Information Science, Southwest University, Chongqing, China*
[3]*School of Computer Science and Technology,*
*Dongguan University of Technology, Dongguan, China*
*\* liuzhigangx@gmail.com*

Abstract

**Due to the insufficient diagnosis and treatment capabilities at the grassroots level, there are still deficiencies in the early identification and early warning of acute exacerbation of Chronic obstructive pulmonary disease (COPD), often resulting in a high prevalence rate and high burden, but the screening rate is relatively low. In order to gradually improve this situation. In this paper, this study develop a Causal Heterogeneous Graph Representation Learning (CHGRL) method for COPD comorbidity risk prediction method that: a) constructing a heterogeneous Our dataset includes the interaction between patients and diseases; b) A cause-aware heterogeneous graph learning architecture has been constructed, combining causal inference mechanisms with heterogeneous graph learning, which can support heterogeneous graph causal learning for different types of relationships; and c) Incorporate the causal loss function in the model design, and add counterfactual reasoning learning loss and causal regularization loss on the basis of the cross-entropy classification loss. We evaluate our method and compare its performance with strong GNN baselines. Following experimental evaluation, the proposed model demonstrates high detection accuracy.**

## I. INTRODUCTION

Chronic obstructive pulmonary disease (COPD) is associated with hypertension may causes even greater harm. Explore the comorbidity intervention strategies under the common pathological mechanism. These comorbidities have a heavy medical and economic burden [1, 2]. Over the past decade, significant progress has been made in the interdisciplinary research of medicine and engineering on comorbidity networks [3]. Some scholars have proposed the method of latent feature representation to handle incomplete high-dimensional feature representations and have achieved good results [4-14]. Some scholars have also discovered new clusters by using matrix methods for community detection [15-45].

Researchers have proposed a wide variety of heterogeneous graph learning classification prediction methods, including traditional heterogeneous graph attention, heterogeneous graph convolution methods, heterogeneous knowledge graph. For instance, classification prediction based on heterogeneous graph neural networks remains a research hot topic. Although the classification prediction method of heterogeneous graph neural networks has demonstrated its advantages, it is difficult to trace the key relationship paths, predict and identify the key influence relationships, and lacks interpretability. Most of the current classification predictions based on heterogeneous graph neural networks are used for drug-target predictions [46]. Some scholars have also proposed methods to improve the message-passing mechanism for homogeneous graph learning, and have made significant progress [47-53]. However, heterogeneous comorbidity networks often have non-Euclidean, nonlinear and incomplete features. Classification predictions based on heterogeneous graph neural networks often have limited effects.

Existing methods can roughly be divided into two types: the first is to design methods based on HAN and HAT. For instance, a model is proposed for the classification prediction based on a heterogeneous graph neural network, however, its classification accuracy is far from satisfactory. The second is the introduction of the Relational Graph Convolutional Network (R-GCN) and its application to two standard knowledge bases to complete tasks: RGCN is related to a recent class of neural networks running on graphs and is specifically developed to handle the highly multi-relational data features of real knowledge bases. For instance, scholars proposed using graph convolutional networks to model relational data.

Causal learning has demonstrated significant advantages in graph representation learning [54-62]. Inspired by this, this studies the multi-relation knowledge graph representation method of causal learning, extracting and fusing patient and network features to achieve higher classification prediction accuracy simultaneously. A comorbidity risk prediction model based on causal attention heterogeneous graph learning predicts the disease status through representation learning of patients' heterogeneous features and relationships.

For the above considerations, we propose a Causal Heterogeneous Graph Representation Learning (CHGRL) prediction method to achieve the purpose of feature fusion of incomplete lab features and comorbidity networks. Main contributions of this paper are:
- Heterogeneous comorbidity datasets and a heterogeneous comorbidity network on a multi-relationship graph;
- A causal graph learning module which automatically learns the causal strength of edge types based on heterogeneous graph neural networks instead of using predefined causal weights; and
- A causal attention mechanism where attention weights have been calculated based on causal strength and causal strength is used to weight message passing for distinguish true causal relationships.

Experiments on heterogeneous comorbidity networks from real-world applications demonstrate that the CHGRL prediction model achieves a significant improvement in accuracy.

## II. RELATED WORK

To discover the comorbidity patterns and predict the risk disease, many methods for comorbidity network analysis and prediction of risk disease have been proposed.

Some research on graph neural networks based on the combination of causal relationships has also been carried out. Many studies have explored the relationship between patient comorbidities and predicted classification by calculating the similarity between patients. Some scholars have also conducted research on causal inference combined with comorbidities. Our CHGRL method overcomes these limitations for better detecting cardiovascular comorbidity.

## III. BACKGROUND

In this section, we introduce the formulation of preparatory knowledge and symbols for the COPD prediction problem. Let the heterogeneous graph be $G=(V, E, T_V, T_E, R)$, where $V$ is the set of patient and disease nodes, $E$ is the set of different edges in the graph, $T_V$ is the set of patient and disease node types, $T_E$ is the set of different edge types in the graph, and $R$ is the set of type relations. For each relation $r \in R$, a relationship-specific adjacency matrix is constructed.

This paper takes the representations of the two homogeneous graph neural networks, patient-patient and disease-disease, as the input of the heterogeneous graph. Our purpose is to predict whether the patient is at risk of developing COPD.

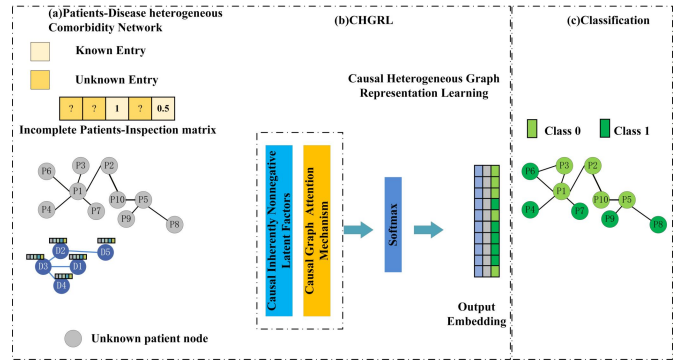

Fig. 1. The framework of a CHGRL framework.

## IV. METHODOLOGY

In this section, we first elaborate on the method of processing of incomplete laboratory data. Then, how to learn the representations of patients, diseases and their interrelationships the proposed risk prediction learning method for COPD comorbidity is introduced. We finally construct the CHGRL method. Fig. 1 illustrates the proposed CHGRL framework.

### A. CSINLF Model for Incomplete Labratory Data

In this study, since the lab features of the patients are incomplete, the latent feature method has been applied to preprocess the missing data values. We establish a bidirectional constraint relationship in the latent feature update process of patient and disease test features, we achieve this goal by introducing the causal regularization term gamma, we proposed a Causal inherently nonnegative latent factors (CSINLF) model.

The motivation is to integrate causal inference into the INLF decomposition process [12-14], combining the robustness of causal relationships with the representational power of INLF, to make latent features more in line with the causal nonnegative latent factor mechanism. After adding the causal regularization term to the objective function, the goal is to learn the U matrix and V matrix that conform to the causal constraints, and the objective function becomes the following formula:

$$\arg\min_{Y} \varepsilon(Y) = \frac{1}{2} \sum_{\forall s_{u,v} \in \Gamma} \left( \left( s_{u,v} - \sum_{k=1}^{d} \delta(y_{(u)k}) \delta(y_{(v)k}) \right)^2 \right. \\ \left. + \mu \sum_{k=1}^{d} \left( \delta^2(y_{(u)k}) + \delta^2(y_{(v)k}) \right) + \gamma \sum_{k=1}^{d} \delta(y_{(u)k}, y_{(v)k}) \right). \quad (1)$$

where $\gamma$ is the regularization intensity parameter, which controls the importance of the causal constraints. $\gamma(\cdot)$ denotes the causal regularization term, punishing the decomposition results that violate the prior causal relationship.

*B. CHGRL Method*

To learn the representations of patients and diseases and their interrelationships simultaneously, it is necessary to model them as heterogeneous graphs. We have developed a method to integrate the representations obtained by the above method CSINLF as the patient feature input of the heterogeneous graph, thereby achieving the purpose of heterogeneous feature fusion.

The convolution formula for the *l*-layer heterogeneous graph is as follows:

$$h_i^l = \sigma \left( \sum_{r \in R} \sum_{j \in N_i^r} W_r h_j^{(0)} \right), \quad (2)$$

where *r* represents all relationship types, and *i* and *j* are neighbor nodes.

The cause-and-effect graph learning formula is as follows:

$$CS = \sigma \left( W_2 \cdot \text{Re} LU \left( W_1 \cdot (h_i \| h_j) + b_1 \right) + b_2 \right). \quad (3)$$

Here, CS represents causal strength, which is used to measure the strength of a certain causal relationship. $\sigma$ is the Sigmoid activation function.

The formula for distinguishing causal relationships of the causal attention mechanism is as follows:

$$\alpha_{ij}^r = \sigma \left( W_4 \cdot \text{Re} LU \left( W_3 \cdot (h_i \| h_j \| CS) + b_3 \right) + b_4 \right), \quad (4)$$

where $\alpha_{ij}^r$ represents the attention weight. *W* denotes the weight matrix, and *h* is the feature matrix.

The intervention formula for calculating counterfactual loss node features is as follows:

$$h_i^{in} = W_6 \cdot \text{Re} LU \left( W_6 \cdot \text{Re} LU \cdot (W_6 \cdot h_i) + b_5 \right) + b_6, \quad (5)$$

where $h_i^{in}$ is the feature vector after intervention, used for tasks related to causality and intervention, and b is the bias term.

The formula for causal message propagation is as follows:

$$m_{j \leftarrow i} = \alpha_{ij} \cdot h_i, \quad (6)$$

where $m_{j \leftarrow i}$ represents the causal message between nodes *i* and *j*, and $\alpha_{ij}$ represents the causal attention weight.

The message aggregation formula is as follows:

$$h_i^{cs} = \sum_{i \in N(j)} m_{j \leftarrow i}, \quad (7)$$

where $h_i^{cs}$ is the causal feature vector of the aggregated node *j*, *N(j)* representing the neighbor set of node *j*.

The final feature representation formula based on feature fusion is as follows:

$$h_i^f = h_j + h_j^{cs}, \quad (8)$$

where $h_i^f$ is the final feature representation of node *j* and $h_j^{cs}$ is the feature vector of node *j* after causal attention weighting.

The overall objective function is as follows:

$$L_t = L_m + L_{cf} + \lambda L_{cr}. \quad (9)$$

Here, $L_c$ is the total loss function, $L_m$ is the main classification loss function, $L_{cf}$ is the counterfactual loss function, $\lambda$ is the causal regularization coefficient, and $L_{cr}$ is causal regularization.

## V. RESULTS

In this section, we introduce the setup of experiments and validate the effectiveness by CHGRL of COPD comorbidity risk prediction.

### A. Datasets and Experimental Settings

To predict the risk of disease, two groups of patients have been selected. The dataset contains 516 patient nodes and 9256 edges between patients, including 201 in the case group and 315 in the control group. There are 68 patient lab features, 386 disease nodes, and 5299 edges before the disease.

The experiments have been conducted on a PC with a 2.5 GHz i9 CPU and 32GB RAM. Hyperparameters are the optimal values obtained through a large number of experiments. We run each baseline five times to obtain the average performance.

### B. Comparison Baselines

To illustrate the effectiveness of CHGRL, the comparison experiments have been conducted in Table I. We compare the performance of the CHGRL model with the strong HGNN baselines: SHGN [54], HGIN [55], HSAGE [56], RGAT [57], RGCN [58]. We use official codes for all baseline HGNN models, and adopt area under the curve (AUC), Accuracy (ACC) and F1 score as evaluation metrics.

Existing heterogeneous methods seldom take into account the particularity and modelling of heterogeneous comorbidity network data, and the prediction classification has not achieved the desired effect. Moreover, they do not simultaneously consider the processing of incomplete examination data of patients, nor do they take into account the interpretability.

### C. Classification Performance Comparisons

In this part of experiments, we evaluate CHGRL and compare its performance with several HGNN baselines. Note that $\lambda$, $\eta$ and $hl$ of CHGRL are 0.005, 0.02, and 64, respectively. The average classification performance evaluated by AUC, ACC, F1-weighted has been listed in Table 2. When evaluated by AUC, CHGRL achieves the highest accuracy at 0.8339, which is 5.56% higher than 0.78 by RGCN, 12.46 % higher than 0.7415 by SHGN, 7.41 % higher than 0.7764 by HGIN, 10.38 % higher than 0.7555 by HSAGE, 14.31% higher than 0.7295 by RGAT. When evaluated by ACC, CHGRL achieves the highest accuracy at 0.8165, which is 4.99% higher than 0.7777 by RGCN, 6.83 % higher than 0.7643 by SHGN, 5.94 % higher than 0.7707 by HGIN, 5.50 % higher than 0.7739 by HSAGE, 13.3 % higher than 0.7206 by RGAT. When evaluated by F1, CHGRL achieves the highest accuracy at 0.809, which is 4.67% higher than 0.7729 by RGCN, 7.64 % higher than 0.7516 by SHGN, 6.62% higher than 0.7588 by HGIN, 7.14 % higher than 0.7551 by HSAGE, 13.62 % higher than 0.712 by RGAT.

We can find from Fig. 2 that the patients in the case group and the control group achieved good classification results after causal representation learning, indicating that our method is suitable for the scenario of differentiating patients with heterogeneous comorbidities. The classification effect and interpretability of the model have been improved based on causal reasoning.

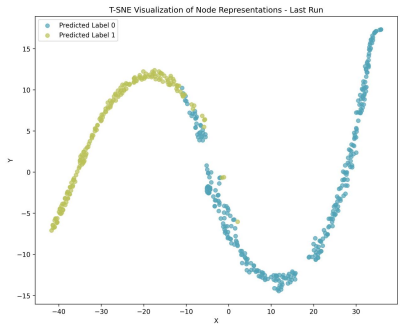

Fig. 2. The classification effects of the two groups of patients by CHGRL.

TABLE I. COMPARISON RESULTS OF CLASSIFICATION EFFECTS WITH DIFFERENT PREDICTION METHODS BY FIVE ROUNDS OF COPD.

| Method | AUC | ACC | F1 |
|---|---|---|---|
| SHGN | $0.7415_{\pm 0.0360}$ | $0.7643_{\pm 0.0270}$ | $0.7516_{\pm 0.0260}$ |
| HGIN | $0.7764_{\pm 0.0454}$ | $0.7707_{\pm 0.0581}$ | $0.7588_{\pm 0.0686}$ |
| HSAGE | $0.7555_{\pm 0.0410}$ | $0.7739_{\pm 0.0478}$ | $0.7551_{\pm 0.0478}$ |
| RGAT | $0.7295_{\pm 0.0617}$ | $0.7206_{\pm 0.0473}$ | $0.7120_{\pm 0.0444}$ |
| RGCN | $0.7900_{\pm 0.0356}$ | $0.7777_{\pm 0.0491}$ | $0.7729_{\pm 0.0472}$ |
| **CHGRL** | $\mathbf{0.8339}_{\pm 0.0455}$ | $\mathbf{0.8165}_{\pm 0.0290}$ | $\mathbf{0.8090}_{\pm 0.0331}$ |

## VI. Conclusion

This study proposes a novel CHGRL method for comorbidity risk prediction by detecting COPD comorbidity. To do this, we combine constraints with latent factor interventions to form a dual-path guidance for causal relationships, which applies to the clinical prediction task of modeling causal associations. This is the first study to conduct the classification of the comorbidity network, however, there are still some limitations call for our future efforts. Our future research will focus on the Granger causality test and integrate multi-modal data to enhance prediction accuracy. Moreover, in addition to the coefficients learned in the joint attention computation, we will also explore more coefficients calculated in heterogeneous patient graphs and time series graph data.

## Acknowledgment

The work is partly funded by Guangdong Basic and Applied Basic Research Foundation under the Grant 2023A1515110689, and partly funded by China Postdoctoral Science Foundation under the Grant 2025M770780.